\documentclass[letterpaper, 10 pt, conference]{ieeeconf}  

\IEEEoverridecommandlockouts                              

\usepackage{amsmath}
\usepackage{xcolor, soul}
\usepackage{subcaption}
\usepackage{amsmath}
\usepackage{amsfonts}
\usepackage{amssymb}
\usepackage{bm}
\usepackage{multicol}
\usepackage{multirow}
\usepackage{optidef}
\usepackage{scalerel}
\usepackage{tikz}
\usetikzlibrary{svg.path}
\usepackage{colortbl}
\usepackage{color}
\definecolor{celadon}{rgb}{0.67, 0.88, 0.69}
\usepackage{algorithm}
\usepackage{algpseudocode}
\usepackage{graphicx}
\usepackage{textcomp}
\usepackage{hyperref}
\usepackage{tabularx}
\newcommand\BibTeX{{\rmfamily B\kern-.05em \textsc{i\kern-.025em b}\kern-.08em
T\kern-.1667em\lower.7ex\hbox{E}\kern-.125emX}}
\definecolor{orcidlogocol}{HTML}{A6CE39}
\tikzset{
  orcidlogo/.pic={
    \fill[orcidlogocol] svg{M256,128c0,70.7-57.3,128-128,128C57.3,256,0,198.7,0,128C0,57.3,57.3,0,128,0C198.7,0,256,57.3,256,128z};
    \fill[white] svg{M86.3,186.2H70.9V79.1h15.4v48.4V186.2z}
                 svg{M108.9,79.1h41.6c39.6,0,57,28.3,57,53.6c0,27.5-21.5,53.6-56.8,53.6h-41.8V79.1z M124.3,172.4h24.5c34.9,0,42.9-26.5,42.9-39.7c0-21.5-13.7-39.7-43.7-39.7h-23.7V172.4z}
                 svg{M88.7,56.8c0,5.5-4.5,10.1-10.1,10.1c-5.6,0-10.1-4.6-10.1-10.1c0-5.6,4.5-10.1,10.1-10.1C84.2,46.7,88.7,51.3,88.7,56.8z};
  }
}

\newcommand\orcidiconKLW[1]{\href{https://orcid.org/0000-0002-1938-4222}{\mbox{\scalerel*{
\begin{tikzpicture}[yscale=-1,transform shape]
\pic{orcidlogo};
\end{tikzpicture}
}{|}}}}

\newcommand\orcidiconAJP[1]{\href{https://orcid.org/0000-0002-8864-9044}{\mbox{\scalerel*{
\begin{tikzpicture}[yscale=-1,transform shape]
\pic{orcidlogo};
\end{tikzpicture}
}{|}}}}

\newcommand\orcidiconHYC[1]{\href{https://orcid.org/0000-0003-1074-4225}{\mbox{\scalerel*{
\begin{tikzpicture}[yscale=-1,transform shape]
\pic{orcidlogo};
\end{tikzpicture}
}{|}}}}
\newcommand\orcidiconFGS[1]{\href{https://orcid.org/0000-0002-5090-9007}{\mbox{\scalerel*{
\begin{tikzpicture}[yscale=-1,transform shape]
\pic{orcidlogo};
\end{tikzpicture}
}{|}}}}
\newcommand\orcidiconAAS[1]{\href{https://orcid.org/0000-0002-6140-619X}{\mbox{\scalerel*{
\begin{tikzpicture}[yscale=-1,transform shape]
\pic{orcidlogo};
\end{tikzpicture}
}{|}}}}
\newcommand\orcidiconLCS[1]{\href{https://orcid.org/0000-0002-9184-3710}{\mbox{\scalerel*{
\begin{tikzpicture}[yscale=-1,transform shape]
\pic{orcidlogo};
\end{tikzpicture}
}{|}}}}

\title{\LARGE \bf
Closed-Loop Control and Disturbance Mitigation of an Underwater Multi-Segment Continuum Manipulator}

\author{Kyle L. Walker$^{1,4 \orcidiconKLW{0000-0002-1938-4222}}$, Hsing-Yu Chen$^{1 \orcidiconHYC{0000-0003-1074-4225}}$, Alix J. Partridge$^{1 \orcidiconAJP{0000-0002-8864-9044}}$, Adam A. Stokes$^{2 \orcidiconAAS{0000-0002-6140-619X}}$, \\ Lucas Cruz da Silva$^{3 \orcidiconLCS{0000-0002-9184-3710}}$ and Francesco Giorgio-Serchi$^{2 \orcidiconFGS{0000-0002-5090-9007}}$
\thanks{This work was supported by Brazilian Agency for Industrial Research and Technological Innovation (EMBRAPII).}
\thanks{$^{1}$Kyle L. Walker, Hsing-Yu Chen and Alix J. Partridge are with the National Robotarium, Boundary Road North, Heriot Watt University Campus, Edinburgh, U.K.}%
\thanks{$^{2}$Adam A. Stokes and Francesco Giorgio-Serchi is with the Institute for Integrated Micro and Nano Systems, University of Edinburgh, Edinburgh, U.K. Correspondence: {\tt\small f.giorgio-serchi@ed.ac.uk}.} %
\thanks{$^{3}$Lucas Cruz da Silva is with the Robotics Department, Senai Cimatec, Salvador, Bahia, Brazil.}%
\thanks{$^{4}$Kyle L. Walker is also with the CREATE Lab, EPFL, Lausanne, Vaud, Switzerland.}%
}

\begin{document}

\maketitle
\thispagestyle{empty}
\pagestyle{empty}

\begin{abstract}


The use of soft and compliant manipulators in marine environments represents a promising paradigm shift for subsea inspection, with devices better suited to tasks owing to their ability to safely conform to items during contact. However, limitations driven by material characteristics often restrict the reach of such devices, with the complexity of obtaining state estimations making control non-trivial. Here, a detailed analysis of a 1m long compliant manipulator prototype for subsea inspection tasks is presented, including its mechanical design, state estimation technique, closed-loop control strategies, and experimental performance evaluation in underwater conditions. Results indicate that both the configuration-space and task-space controllers implemented are capable of positioning the end effector to desired locations, with deviations of $<$5\% of the manipulator length spatially and to within 5$^{o}$ of the desired configuration angles. The manipulator was also tested when subjected to various disturbances, such as loads of up to 300g and random point disturbances, and was proven to be able to limit displacement and restore the desired configuration. This work is a significant step towards the implementation of compliant manipulators in real-world subsea environments, proving their potential as an alternative to classical rigid-link designs.




\end{abstract}

\section{Introduction}
\label{sec: intro}



As energy demands continue to rise across the globe \cite{WANG2023101048}, so does the need for increased and well maintained infrastructure. As such, maintaining current pipelines is vital to ensuring demands are met \cite{Ho2020}, while also reducing the impact to marine environments \cite{KOPPEL2023107093}. Due to the inaccessibility of subsea pipelines for engineers, robotic solutions have been developed for inspection, with ROVs being an established technology and a push in recent years towards AUVs \cite{zhang2019, WalkerICRA}. A key benefit of such platforms is their ability to carry tailored sensor payloads to inspection sites, including imaging \cite{WANG2023462}, acoustic \cite{YU2021} and magnetic sensors \cite{peng2020}, providing a broad picture of the health of the inspected asset \cite{Mai2016}. In addition, ROVs and AUVs provide a strong base for the addition of manipulators that can be used for both inspection and maintenance \cite{SIVCEV2018431}. 

While traditional manipulators enable a wide range of additional abilities, for example intervention \cite{Christensen2022} or cleaning tasks \cite{zapico2024}, this is typically met with a trade-off between weight and effective reach, restricting the use of specific devices to specific vehicles \cite{SIVCEV2018431}. In turbulent environments, this can prevent contact-inspection and similar tasks using small-scale ROVs, as the length of manipulator the ROV can carry as a payload is fundamentally limited, thus the risk of the ROV inadvertently colliding with the structure increases \cite{WalkerIJRR, WalkerIROS}. Likewise, the lack of flexibility and compliance within the rigid links of the manipulator mean additional advanced control is necessary to avoid damage to the plant (or the vehicle) \cite{zapico2024, Tugal2023}; on a floating base this complexity is exacerbated.

Advances in the field of continuum and soft robotics have displayed potential for low weight to high length devices \cite{Tonapi2015}, with the additional benefit of inherent structural compliance increasing resilience within dynamic subsea environments \cite{Qu2024Recent}. In contrast to typical rigid-link designs, soft manipulators (or manipulators exhibiting traits associated with soft robotics) are often pneumatically actuated \cite{Liu2020}, a characteristic undesirable for underwater operation. Tendons are another common actuation method, but often designs are constrained to shorter length scales \cite{Nguyen2015, Wockenfub2022} and limited in terms of operation under payload \cite{Xie2020}, due to their reduced load-bearing capacity as a product of their increased flexibility. Compliant manipulators have recently shown promise at achieving increased reach and load-bearing capabilities in terrestrial environments \cite{guan2023trimmed}, with some devices exploiting variable stiffness to grant manipulators benefits from both compliant and traditional devices \cite{mukaide2020radial}. However, for instances where larger length-scales have been achieved, motion is typically constrained \cite{Wang2021, Wang2018} with a focus on a specific task/scenario \cite{Dong2017}.

\begin{figure}[t!]
    \centering
    \includegraphics[width=\linewidth]{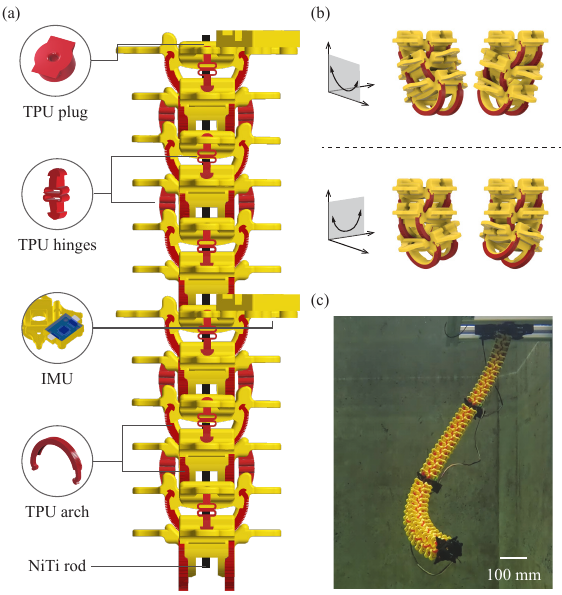}
    \caption{(a) Diagram of the assembled manipulator, showing: a TPU plug, secured at the bead centre for routing a 2mm diameter NiTi rod; TPU arches between beads; an IMU encapsulated in resin; compressible TPU hinges. Also, (b) an example of how neighbouring beads move in orthogonal planes and (c) the manipulator in operation underwater.}
    \label{fig: design}
\end{figure}

This work focuses on these challenges, building on previous research into modular, tendon driven, variable stiffness compliant manipulators \cite{walker2024modular}. Significant advancements are demonstrated through closed-loop set-point regulation and disturbance compensation for deployment in realistic subsea environments (Fig. \ref{fig: design}). The remainder of the paper is structured as follows: Section \ref{sec: design_modelling} details the mechanical design, sensor integration and kinematics of the manipulator; Section \ref{sec: control_methods} details the control methods, including state estimation, closed-loop control and tension supervision; Section \ref{sec: experiments} presents experimental setup and testing of the manipulator, encapsulating underwater configuration-space posture tracking, task-space end-effector tracking and disturbance compensation before giving concluding remarks.

\section{Design \& Modelling}
\label{sec: design_modelling}
\subsection{Mechanical Design}
\label{subsec: mechanical_design}

The manipulator demonstrated here extends prior work into modular, tendon driven, variable stiffness manipulators \cite{walker2024modular} as follows: the manipulator length is increased $\sim$2.5x and has an additional controllable segment; a method of state estimation is integrated; closed-loop control is validated experimentally. Furthermore, operation of the manipulator is proven in an underwater environment. The manipulator is fully modular and can be assembled to various lengths, with varying number of segments. For this work, a manipulator 1m in length with three independently controllable segments is presented. 

Modularity is achieved by utilising repeating pieces that slot together until a desired length is reached. In this design the modular pieces are as follows: PLA beads; compressible TPU hinges; TPU arches; TPU plugs; a 2mm diameter NiTi rod equal in length to the desired length of the manipulator. These pieces are presented in Fig. \ref{fig: design}. Each bead is 29.5mm in height, 62mm wide and weighs 10g, with 1 segment of the manipulator comprising 16 beads (15 regular beads and 1 with the addition of a holster for a waterproofed IMU). Another addition to the design is the inclusion of a 2mm diameter nitinol rod down the centre of the manipulator that acts as a sprung backbone. To hold the rod in place, TPU plugs with a central hole have been added to the centre of each bead that guide the rod, while allowing beads to compress and adding the potential for stiffening. 

Further advancing on \cite{walker2024modular}, beads slide over arched \textit{rails} manufactured in TPU, that increase static friction between successive beads during tensile loading of the manipulator backbone. Successive beads rotate around respectively orthogonal axes, as shown in Fig. \ref{fig: design}(b). Each section of the manipulator is controlled by 4 tendons, with each tendon connecting via bowden cable to a single servo motor (Dynamixel, XM430-W350). The assembled manipulator in operation is displayed in Fig. \ref{fig: design}(c).

 \begin{figure}[t!]
    \centering
    \includegraphics[width=\linewidth]{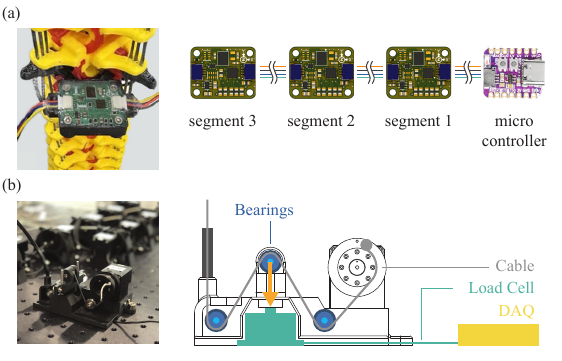}
    \caption{Overview of the sensing system implemented within the manipulator design. (a) IMUs are daisy-chained to estimate the shape of the manipulator and (b) load cells are integrated within the motor housing to provide cable tension feedback.}
    \label{fig: sensors}
\end{figure}

\subsection{Sensor Integration}
\label{subsec: sensor_integration}

To accurately reconstruct the configuration of the manipulator, an Inertial Measurement Unit (IMU) is positioned at the end of each segment. The selected IMU (Bosch, BNO055) is a 3-axis gyro, 3-axis magnetometer and 3-axis accelerometer sensor providing absolute orientation measurements in quaternion format. Given the BNO055 has only two configurable I2C addresses, a multiplexer was incorporated to facilitate additional connections. Our custom-designed Printed Circuit Board (PCB) integrates the IMUs with a multiplexer (Texas Instruments, TCA9548A) , enabling daisy-chaining of up to eight sensors, which can then be connected to a micro-controller. To ensure waterproofing, the entire IMU assembly is encapsulated in epoxy resin within a PLA holster on the bead (see Fig. \ref{fig: design}). The design for the assembly is open-source and freely available \textcolor{blue}{\href{https://github.com/azurechen1203/multiple-multiplexer-BNO055-interface}{here}).}

Monitoring cable tension in a tendon-driven system is critical for providing feedback on the system state, maintaining desired performance, and enhancing operational reliability. Excessive tension can cause cables to snap, while insufficient tension can lead to slack cables, both of which can result in unpredictable movements or damage to the manipulator. Here, a tension monitoring system is implemented by integrating a motor, a load cell (TE Connectivity, FC2231), and three pulleys for each cable (see Fig. \ref{fig: sensors}). This design ensures that the forces acting along the cable remain co-planar, so that as the cable experiences tension, it compresses the load cell which converts this downward force into an electrical signal, acquired by a LabJack data acquisition system (DAQ) . The transformation to a tension reading is performed according to the manufacturer data sheet and a pre-experiment characterisation, facilitating the tension supervision scheme described in Section \ref{subsec: tension_comp}. It is worth noting that this setup is out of water for characterisation purposes.




\subsection{Kinematics}
\label{subsec: kinematics}
\begin{figure}[t!]
    \centering
    \includegraphics[width=0.9\linewidth]{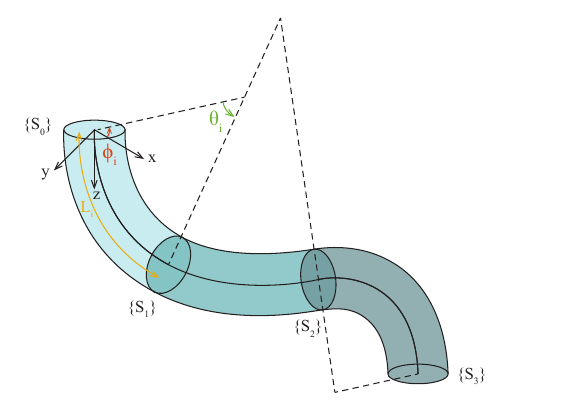}
    \caption{Piecewise Constant Curvature (PCC) kinematic model of a single spatial segment, where the dashed lines represent the bending plane.}
    \label{fig: pcc_kinematics}
\end{figure}

The manipulator consists of three independently actuated segments, each driven by a set of four tendons. Each segment is assumed to follow a Piecewise Constant Curvature (PCC) approximation \cite{Webster_Jones_2010, Rucker_Webster_III_2011} (see Fig. \ref{fig: pcc_kinematics}). The interlocking interface and hinges between each bead resist twisting, whilst shortening is assumed negligible ($<$1\%) due to the low compressibility of the TPU arches. 

Consider a reference frame attached to the end of each segment $\{S_i\}$, with an additional frame attached to the base $\{S_0\}$. For the $i$-th segment, the configuration variable consists of the bending plane angle, $\phi_i$, and curvature angle, $\theta_i$, such that $q_i = [\phi_i, \theta_i]^T \in \mathbb{R}^2$; the complete configuration of the manipulator in this work is therefore defined as $q = [ q_1, q_2, q_3 ]^T \in \mathbb{R}^6$ . For each segment (from base to tip) the translation mapping from $\{S_{i-1}\}$ is denoted as $t_{i-1}^{i} \in \mathbb{R}^3$ and defined as:
\begin{equation}
    t_{i-1}^{i}(\phi_i, \theta_i, L_i) =  \frac{L_i}{\theta_i} \begin{bmatrix}
        \cos\phi_i(1-\cos\theta_i) \\
        \sin\phi_i(1-\cos\theta_i) \\
        \sin\theta_i
    \end{bmatrix}
\end{equation}
whilst the orientation mapping is denoted $R_{i-1}^{i} \in \mathbb{R}^3$ and defined as:
\begin{align}
    & R_{i-1}^{i}(\phi_i, \theta_i) = \nonumber\\
    & \begin{bmatrix}
        c^{2}_{\phi_i}(c_{\theta_i}-1)+1 & s_{\phi}c_{\phi}(c_{\theta}-1) & s_{\phi}c_{\theta} \\
        s_{\phi}c_{\phi}(c_{\theta}-1) & c^{2}_{\phi_i}(1-c_{\theta_i})+c_{\theta_i} & s_{\phi}s_{\theta} \\
        -c_{\phi}s_{\theta} & -s_{\phi}s_{\theta} & c_{\theta_i}
    \end{bmatrix} 
\end{align}
where $L_{i}$ is the segment length and $c(\cdot), s(\cdot)$ are shorthand for $\cos(\cdot), \sin(\cdot)$. Finally, the homogeneous transformation matrix between frames with respect to the base frame of the segment is defined as:
\begin{equation}
    T_{i}(\phi_i, \theta_i) = \begin{bmatrix}
        R_{i-1}^{i}(\phi_i, \theta_i) & t_{i-1}^{i}(\phi_i, \theta_i) \\
        \mathbf{0}_{1\times3} & 1
    \end{bmatrix}
\end{equation}
This facilitates deduction of the manipulator end-effector position through knowledge of the configuration variable, extracted via the integrated IMUs (see Section \ref{subsec: state_estimation}).



\begin{figure}[t!]
    \centering
    \includegraphics[width=0.99\linewidth]{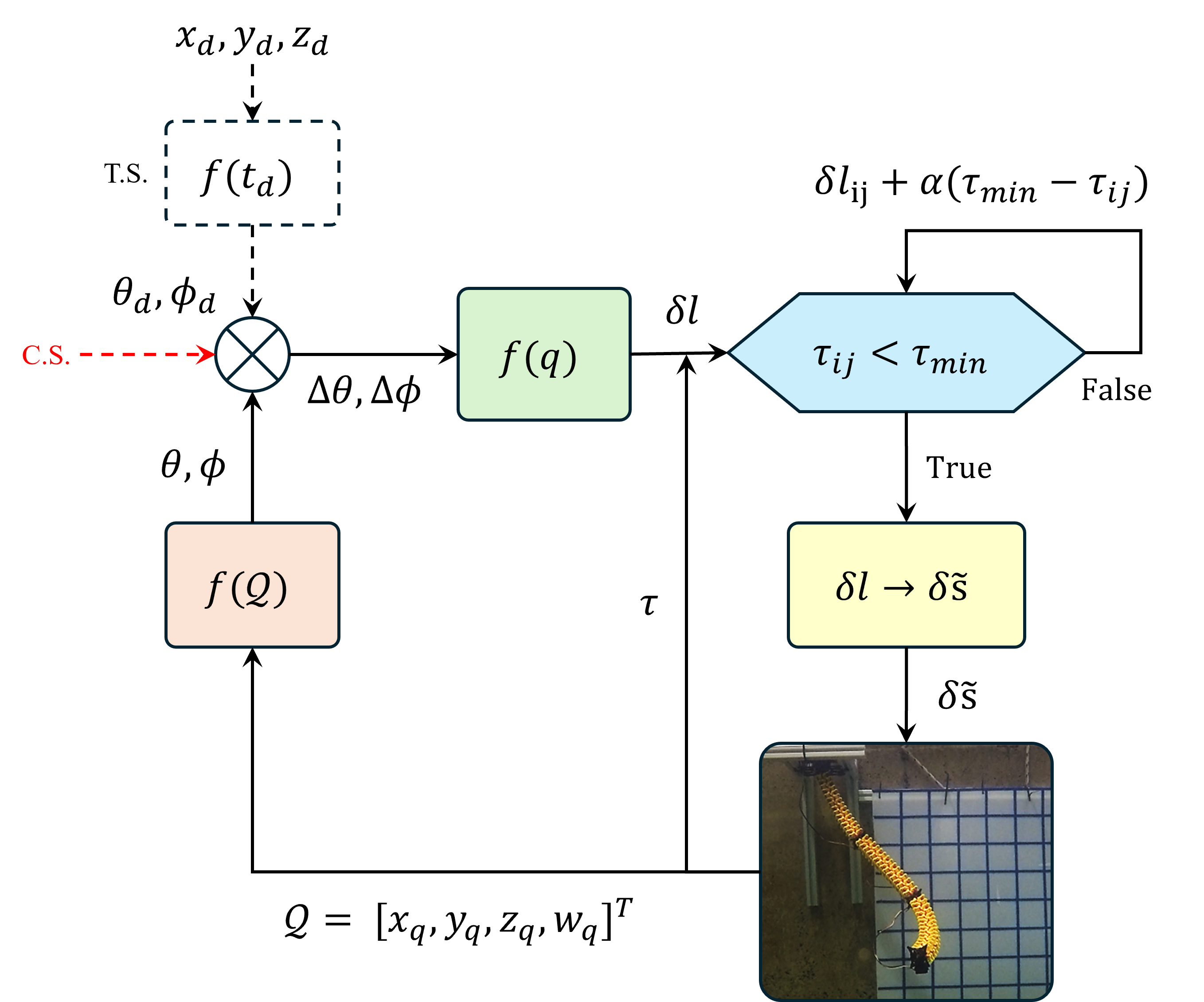}
    \caption{Block diagram of the control architecture, where the dashed elements indicate procedures only in use during task-space feedback control.}
    \label{fig: control_bd}
\end{figure}

\section{Control Methodology}
\label{sec: control_methods}
\subsection{State Estimation}
\label{subsec: state_estimation}
To estimate the state of the robot using the PCC model definition, the arc parameters of each segment must be inferred. As previously described, an IMU placed at the tip of each segment is used to directly measure the orientation in quaternions, $\mathcal{Q} = [x_q, y_q, z_q, w_q]^{T}$ relative to the sensor's fixed-frame, as shown in Fig. \ref{fig: pcc_kinematics}. It follows that the bending plane angle and curvature angle can be deduced through:
\begin{equation}
    \phi_{i} = \arctan\left( \frac{x_q z_q-w_q y_q}{y_q z_q+w_q x_q}\right)
\end{equation}
\begin{equation}
    \theta_{i} = \arccos(2w_q^2-1+2z_q^2)
\end{equation}
The configuration variable $q_{i}$ in the manipulator local-frame is then induced accordingly using the manipulator base as a reference.
The above operations are denoted by the function $f(\mathcal{Q})$ in Fig. \ref{fig: control_bd}.
\subsection{Closed-loop Feedback Control}
\label{subsec: closed_loop_control}
With the state information, formulating a closed-loop controller is based on the assumption a map exists between the actuator space and the configuration space, $l = f(q)$ where $l \in \mathbb{R}^{3\times 4} $ as each segment is actuated by 4 tendons. As the manipulator has multiple segments, this map is computed by defining the tendon lengths as:
\begin{equation} \label{eq: tendon_length}
    l_{ij} = \sum^{i}_{1} l_{b,i} - r_{t}\theta_{i}\cos\left( \frac{(2j-1)\pi}{n_t} - \phi_i \right)
\end{equation}
where $i$ and $j$ refer to the segment and tendon index respectively, $l_{b,i}$ is the segment backbone length and $r_t$ is the radius of the tendon location from the segment backbone. Also, $j \in [1, 2, \dots, n_t]$ where $n_t$ is the number of tendons per segment.

Furthermore, the set of tendon length changes to drive the manipulator to a desired configuration, $q_d = [ \phi_d, \theta_d ]^T$, considers the Jacobian of this mapping with respect to the configuration variable, $\mathcal{J}_l = \frac{\partial l}{\partial q} \in \mathbb{R}^{6\times 6}$. This facilitates formulation of a closed-loop configuration-space control law:
\begin{equation} \label{eq: config_space_control_law}
    \delta l = \gamma(\mathcal{J}_l \Delta q)
\end{equation}
where $\gamma$ is a reduction factor and $\Delta q = q_d - q$.

Extending to task-space control requires an additional transformation to calculate a set of admissible configuration variables, before applying the same methodology as above. In this work, a classic numerical method was implemented to find a solution for the inverse kinematics, by providing a desired end-effector position in 3D space, $t_{d} = [x_d, y_d, z_d]^{T} \in \mathbb{R}^3$, and using the current manipulator configuration as an initial guess, $q_0 = q \in \mathbb{R}^6$ \cite{LynchBook}.

The block diagram representation of the above is shown in Fig. \ref{fig: control_bd}. Additionally, an error check is included to account for the boundary condition where $\phi_i=\pm \pi$, ensuring the shortest path is always taken.

\begin{figure}[t!]
    \centering
    \includegraphics[width=\linewidth]{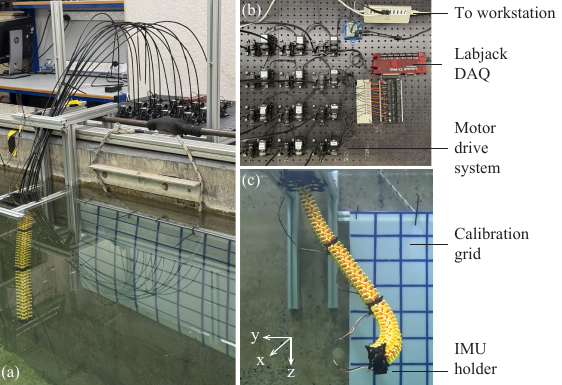}
    \caption{The experimental arrangement, showing (a) the setup of the manipulator and cable routing to the base of the test rig, (b) the motor layout and peripheral connections on the base and (c) the manipulator during operation underwater. }
    \label{fig: exp_setup}
\end{figure}

\subsection{Tension Supervision}
\label{subsec: tension_comp}

Although the tendon lengths and configuration space are related through Eq. \ref{eq: tendon_length}, this does not account for effects such as friction, cable elongation, imperfect posture changes and other un-modelled effects which can lead to slack cables. To account for this, the drive system includes a compressive load cell to directly measure the cable tension and maintain a minimum threshold, $\tau_{min}$; see Fig. \ref{fig: control_bd}. However, as the cables are controlled simultaneously and constantly switched between active and passive, this correction to $\delta l$ has to be performed retrospectively to avoid influencing the posture control negatively. 

The following pseudo-code demonstrates how tension compensation was incorporated in the control procedure:
\begin{equation}\label{eq: tension_compensation}
    \text{if} \quad  \tau_{ij} < \tau_{min}: \quad \delta l_{ij} = \delta l_{ij} + \alpha(\tau_{min}-\tau_{ij}) 
\end{equation}
where $\tau_{ij}$ is the tendon tension and $\alpha$ is a scaling factor. This method of proportional active compensation helps to retain tension quicker the slacker cables become.


\renewcommand{\arraystretch}{1.2}

\begin{table}[t!] 
\vspace*{3mm}
\centering
\caption{Posture specifications for the configuration-space tracking tests, where $\phi_{d}$ is with reference to $\{S_{0}\}$ and $\theta_{d}$ is with reference to $\{S_{i-1}\}$.}
\begin{tabularx}{0.45\textwidth}{|>{\centering\arraybackslash}X|>{\centering\arraybackslash}X|>{\centering\arraybackslash}X|}
\hline
Posture Reference & $\phi_d$ ($^{o}$) & $\theta_d$  ($^{o}$)\\ 
\hline \hline
1 & (-90, -90, -90) & (0, 40, 5) \\ \hline 
2 & (-90, -90, -90) & (40, 10, -5) \\ \hline
3 & (-90, -90, -90) & (40, 10, 35) \\ \hline
4 & (-90, -90, -90) & (40, 10, -40) \\ \hline
5 & (-45, -45, -45) & (40, 10, -5) \\ \hline
6 & (-45, -45, -45) & (40, 10, 35) \\ \hline
7 & (-45, -45, 0) & (40, 10, 35) \\ \hline
8 & (-45, -45, -90) & (40, 10, 35) \\ \hline
\end{tabularx}
\label{tab:des_posture_specs}
\end{table}


\begin{figure}[t!]
    \centering
    \includegraphics[width=\linewidth]{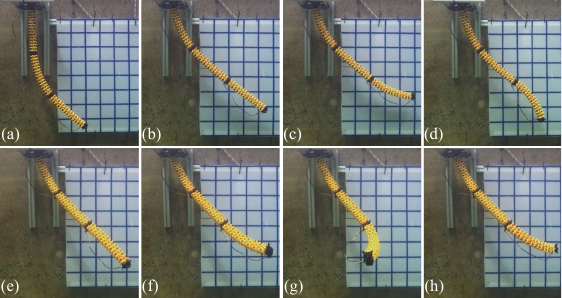}
    \caption{Visual example of each posture listed in Table \ref{tab:des_posture_specs}, in ascending numerical order from (a) to (h).}
    \label{fig: posture_examples}
\end{figure}

\section{Experimental Study}
\label{sec: experiments}
To investigate operation of the manipulator with the autonomous controllers detailed above, several experiments were performed with the manipulator body fully submersed in a water tank and operational. As discussed in Section \ref{subsec: sensor_integration}, the custom PCB was encapsulated in epoxy resin to waterproof the system, whilst the drive system for the manipulator was kept outside of the tank for these initial experiments; see Fig. \ref{fig: exp_setup}. 

To obtain a ground-truth of the end-effector position, a grid was placed in the tank with 100mm spacings between each line (as seen in Fig. \ref{fig: posture_examples}), using Kinovea (a video processing tool) to analyse video footage and obtain positional data relative to the grid spacing and camera perspective. 



\subsection{Configuration-Space Posture Tracking}
\label{subsec: joint_space_tracking}

\begin{figure}[t!]
    \centering
    \includegraphics[width=\linewidth]{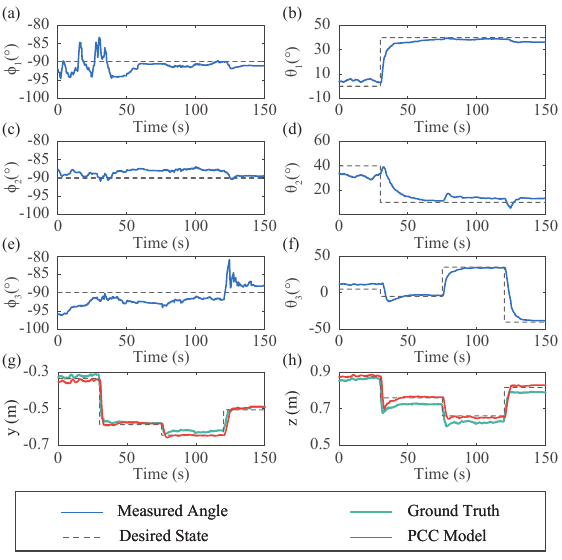}
    \caption{A temporal segment of a configuration-space tracking experiment, showing the evolution of (a)(c)(e) $\phi$ and (b)(d)(f) $\theta$ for each segment $i \in [1, \dots, 3]$ of the manipulator. Also shown are the end-effector positions in the (g) y-axis and (h) z-axis and their accuracy relative to the ground truth obtained via motion tracking.}
    \label{fig: traj_example}
\end{figure}

Evaluating performance in the configuration-space is concerned with the ability of the feedback controller to track a desired posture, $q_d$, in a repeatable/reliable manner. For these experiments, $q_d$ was varied periodically every $45$s according to the pre-defined trajectory given in Table \ref{tab:des_posture_specs}, with visual examples provided in Fig. \ref{fig: posture_examples}. A temporal example of the posture tracking feedback controller operating is given in Fig. \ref{fig: traj_example}, with subplots (a),(c),(d) depicting the time-history of $\theta_{d}, \theta$ and (b)(d)(e) showing $\phi_{d}, \phi$. Also, subplots (g)(h) show the manipulator end-effector position, estimated using the PCC model and using motion tracking software to obtain a ground truth. Here, deviations of $<$5\% of the manipulator length ($\sim$1m) can be witnessed across the trajectory, providing confidence the configuration-space controller is capable of positioning the end-effector at the desired location according to the PCC model.

\begin{figure}[t!]
    \centering
    \includegraphics[width=\linewidth]{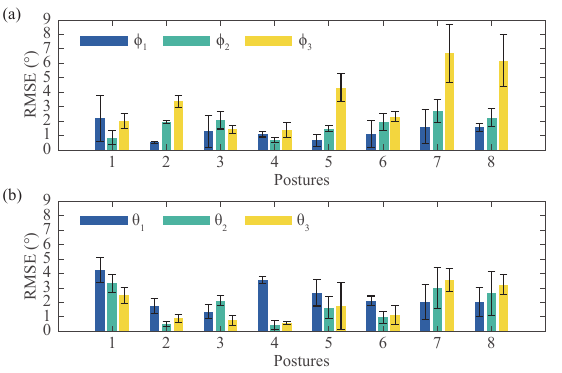}
    \caption{RMSE of the configuration-space tracking experiments, averaged over three trials for each posture. Shown are the errors for (a) $\phi$ and (b) $\theta$ for each segment separately.}
    \label{fig: js_results}
\end{figure}

To analyse the results quantitatively, the control was deemed to have positioned the manipulator to a steady-state posture for the final $5$s of each $45$s window; this period is used to evaluate the tracking accuracy by means of a Root-Mean-Square-Error (RMSE) across three separate trials. With reference to Fig. \ref{fig: js_results}, for both $\phi$ and $\theta$ these results clearly indicate the controller's ability to autonomously regulate the posture with minor errors, remaining within $5^{o}$ for almost all of the postures and all of the trials. Posture 7 and 8 are the only cases where there is a significant increase in error for $\phi_3$, however these are also the cases where segment 3 was commanded to bend in a different plane to the previous two segments. 
Analysing further, the deviation in error between the different trials remains low ($\sim$1-2$^{o}$ for almost all postures), indicating a good degree of repeatability during the manipulator operation.

\begin{figure}[t!]
    \centering
    \includegraphics[width=\linewidth]{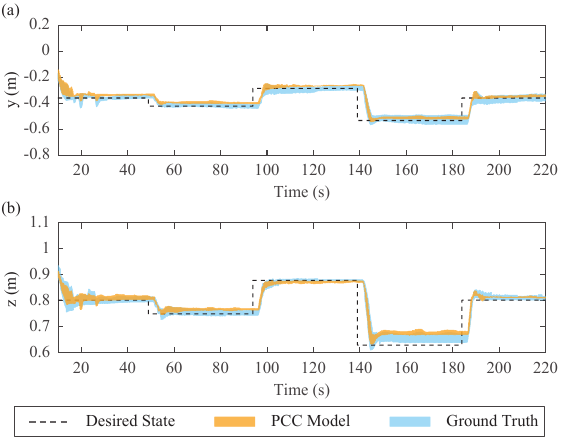}
    \caption{Task-space tracking results, showing the positional tracking of the (a) y-axis and (b) z-axis. As before, a ground-truth obtained via motion tracking is shown for comparison to the PCC approximation.}
    \label{fig: ts_results}
\end{figure}

\subsection{Task-Space End-Effector Tracking}
\label{subsec: task_space_tracking}

For task-space control, an additional layer was added to the control architecture (Fig. \ref{fig: control_bd}) in the form of a numerical inverse kinematics solver, as alluded to in Section \ref{subsec: closed_loop_control}. The focus is now placed on the ability to position the end-effector at a desired spatial coordinate, $t_{d} = [x_d, y_d, z_d]^{T}$, rather than the configuration, $q$. A similar methodology was adopted to that in Section \ref{subsec: joint_space_tracking}, where $t_d$ is varied periodically every 45s according to a pre-defined trajectory and a comparison is drawn against the ground truth. 


As demonstrated by Fig. \ref{fig: ts_results}, the control tracks the desired set-points with remarkable accuracy, displaying errors of $<$3cm for the majority of the experiment, with the exception of the period $150$-$180$s in Fig. \ref{fig: ts_results}(b). Quantitatively, considering the last 5s of each transition as the "steady-state" as before, the average RMSE across all set-points of the end-effector was recorded as 0.026m and 0.015m for the y-axis and z-axis respectively. With reference to the manipulator length, this corresponds to only a 3\% error margin. Coupling this with the manipulators inherent compliance, the implication is that accurate close quarters inspection of target errors could be performed autonomously with confidence. Likewise, given that the controllers implemented for validation are relatively simplistic, this error could be reduced further with effective trajectory planners and advanced control methods \cite{WalkerRoboSoft}. 

\begin{figure}[t!]
    \centering
    \includegraphics[width=\linewidth]{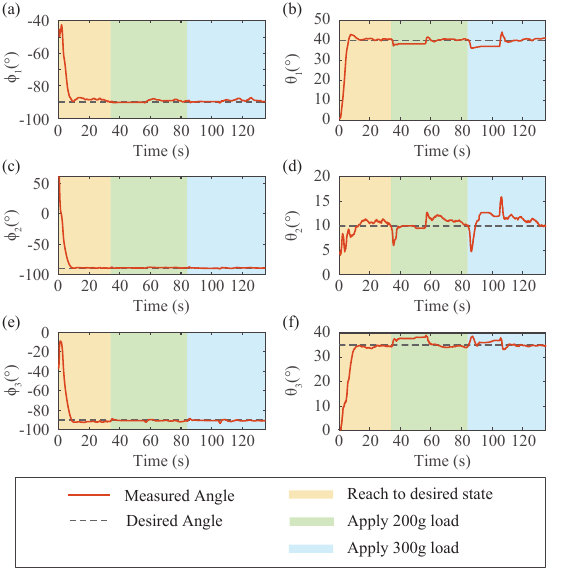}
    \caption{Temporal segment of the weighted disturbance experiments, showing the evolution of (a)(c)(e) $\phi$ and (b)(d)(f) $\theta$ for each segment. Each colour change indicates the point where a different point-load was applied to the end-effector.}
    \label{fig: weighted_dc_results}
\end{figure}

\subsection{Disturbance Mitigation}
\label{subsec: disturbance_compensation}

As the controllers above are purely kinematic and operate on state feedback information in real-time, the ability of the manipulator to react and correct for disturbances to the configuration variables should be inherent. To assess this disturbance rejection performance, the controller was tasked with maintaining a pre-defined set-point in the configuration space, $q_d = [\phi_d, \theta_d]^{T} = [ -90^{o}, 40^{o}, -90^{o}, 10^{o}, -90^{o}, 35^{o} ]^{T},$ during two separate disturbance tests: (1) a point load was applied to the tip by attaching and removing a $200$g and $300$g mass successively during operation, (2) using a rod, random point loads were applied to different locations along the manipulator length. It should be noted that the control methodology has not been altered for these experiments, which are designed to analyse the effect of the manipulator elastic response to a perturbation on the controller performance.

For the tip loading experiment, Fig. \ref{fig: weighted_dc_results} shows the evolution of the manipulator configuration in time, with the beginning of the green and blue shaded areas indicating the points where a 200g and 300g load were applied and released respectively. With reference to Fig. \ref{fig: weighted_dc_results}(a)(c)(e), there is negligible effect on the behaviour of $\phi$ as expected, due to the load being applied vertically. For $\theta$ (Fig. \ref{fig: weighted_dc_results}(b)(d)(f)), it is clear that there is a greater influence on the behaviour of $\theta_1, \theta_3$ than $\theta_2$. For the latter, the control is able to correct the desired configuration even after both of the applied loads have caused deviations. For $\theta_1, \theta_3$, the control was not able to fully correct the posture, however an active attempt to maintain the desired value is still apparent, only deviating by $\sim6^{o}$ from the set-point at most across all variables. 

\begin{figure}[t!]
    \centering
    \includegraphics[width=\linewidth]{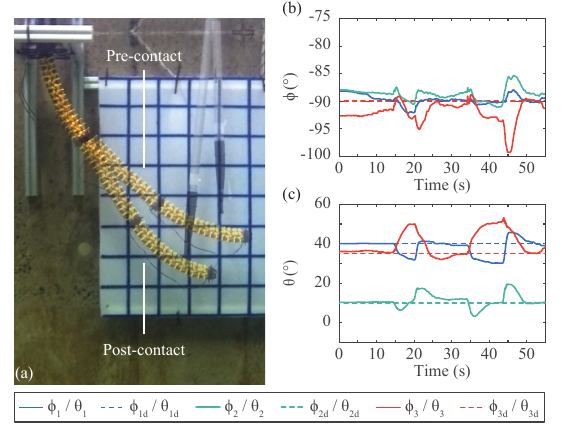}
    \caption{Example of the rod disturbance experiments, showing (a) the manipulator pre and post disturbance evolution, and the evolution of of (b) $\phi$ and (c) $\theta$ for each segment. Each inflection indicates a point of disturbance, after which the controller corrects the manipulator posture.}
    \label{fig: pole_results}
\end{figure}

Moving onto random disturbances using a rod, Fig. \ref{fig: pole_results}(b),(c) display a time history of $\theta$, $\phi$ respectively, with a visual snapshot of the effect on the configuration given in Fig. \ref{fig: pole_results}(a). It is clear that after each instance of disturbance, the feedback controller drives the manipulator back towards the desired set-point. It is visually evident in Fig. \ref{fig: pole_results}(a) that the controller is attempting to retain the desired configuration, as the curvature of segment 3 (where the disturbance is being applied) is visibly higher. Overall these two experiments validate the IMU state estimation and closed-loop control, even in the presence of various disturbances.  

\section{Conclusions}
\label{sec: conclusions}

To the best of our knowledge, this is the first validated implementation of continuous control for a multi-segment manipulator of this length operating spatially in an underwater environment. While improvements can still be made, the controller’s ability to autonomously regulate the posture to within $<$5\% of the manipulators length in both the configuration-space and task-space provides confidence that the control system can position the end-effector to desired locations. Additionally, the deviation in error across trials for configuration tracking was approximately $1-2^{o}$ for almost all postures, indicating that further refinement will result in precise and repeatable control of the manipulator. 

Furthermore, rejection and restoration from different forms of disturbance was demonstrated, specifically 200g/300g loads as well as random point disturbances along the manipulator length. Given the dynamic environment that such a manipulator would be operating in, this gives confidence that unsteady, large magnitude disturbances can be handled. Explicit treatment and inclusion of disturbances within the controller is planned for future work to further improve the system robustness.

Overall, these key experimental findings show the clear potential of large-scale, compliant, underwater manipulators to operate effectively and reliably during subsea operations. With further development, this style of manipulator could offer a solution to the current limitations faced by traditional underwater manipulators, pushing large-scale compliant manipulators towards deployment in real-world environments, for real-world tasks.





\addtolength{\textheight}{-1cm}   








\bibliographystyle{ieeetr}
\bibliography{bibbib}

\begin{thebibliography}{10}

\bibitem{WANG2023101048}
Z.~Wang, S.~Li, Z.~Jin, Z.~Li, Q.~Liu, and K.~Zhang, ``Oil and gas pathway to net-zero: Review and outlook,'' {\em Energy Strategy Reviews}, vol.~45, p.~101048, 2023.

\bibitem{Ho2020}
M.~Ho, S.~El-Borgi, D.~Patil, and G.~Song, ``Inspection and monitoring systems subsea pipelines: A review paper,'' {\em Structural Health Monitoring}, vol.~19, no.~2, pp.~606--645, 2020.

\bibitem{KOPPEL2023107093}
D.~J. Koppel, T.~Cresswell, A.~MacIntosh, R.~{von Hellfeld}, A.~Hastings, and S.~Higgins, ``Threshold values for the protection of marine ecosystems from norm in subsea oil and gas infrastructure,'' {\em Journal of Environmental Radioactivity}, vol.~258, 2023.

\bibitem{zhang2019}
Y.~Zhang, M.~Zheng, C.~An, J.~K. Seo, I.~P. Pasqualino, F.~Lim, and M.~Duan, ``A review of the integrity management of subsea production systems: inspection and monitoring methods,'' {\em Ships and Offshore Structures}, vol.~14, no.~8, pp.~789--803, 2019.

\bibitem{WalkerICRA}
K.~L. Walker, R.~Gabl, S.~Aracri, Y.~Cao, A.~A. Stokes, A.~Kiprakis, and F.~Giorgio-Serchi, ``Experimental validation of wave induced disturbances for predictive station keeping of a remotely operated vehicle,'' {\em IEEE Robotics and Automation Letters}, vol.~6, no.~3, pp.~5421--5428, 2021.

\bibitem{WANG2023462}
H.~Wang, S.~Sun, and P.~Ren, ``Meta underwater camera: A smart protocol for underwater image enhancement,'' {\em ISPRS Journal of Photogrammetry and Remote Sensing}, vol.~195, pp.~462--481, 2023.

\bibitem{YU2021}
Y.~Yu, A.~Safari, X.~Niu, B.~Drinkwater, and K.~V. Horoshenkov, ``Acoustic and ultrasonic techniques for defect detection and condition monitoring in water and sewerage pipes: A review,'' {\em Applied Acoustics}, vol.~183, Dec 2021.

\bibitem{peng2020}
X.~Peng, U.~Anyaoha, Z.~Liu, and K.~Tsukada, ``Analysis of magnetic-flux leakage (mfl) data for pipeline corrosion assessment,'' {\em IEEE Transactions on Magnetics}, vol.~56, no.~6, pp.~1--15, 2020.

\bibitem{Mai2016}
C.~Mai, S.~Pedersen, L.~Hansen, K.~L. Jepsen, and Z.~Yang, ``Subsea infrastructure inspection: A review study,'' in {\em IEEE International Conference on Underwater System Technology: Theory and Applications (USYS)}, pp.~71--76, 13-14 Dec 2016.

\bibitem{SIVCEV2018431}
S.~Sivčev, J.~Coleman, E.~Omerdić, G.~Dooly, and D.~Toal, ``Underwater manipulators: A review,'' {\em Ocean Engineering}, vol.~163, pp.~431--450, 2018.

\bibitem{Christensen2022}
L.~Christensen, J.~Hilljegerdes, M.~Zipper, A.~Kolesnikov, B.~Hülsen, C.~E.~S. Koch, M.~Hildebrandt, and L.~C. Danter, ``The hydrobatic dual-arm intervention auv cuttlefish,'' in {\em OCEANS 2022, Hampton Roads}, pp.~1--8, 17-20 Oct 2022.

\bibitem{zapico2024}
C.~S. Zapico, Y.~Petillot, and M.~S. Erden, ``Semi-autonomous surface-tracking tasks using omnidirectional mobile manipulators,'' in {\em IEEE International Conference on Robotics and Automation (ICRA)}, pp.~2176--2182, 13-17 May 2024.

\bibitem{WalkerIJRR}
K.~L. Walker, L.-B. Jordan, and F.~Giorgio-Serchi, ``Nonlinear model predictive dynamic positioning of a remotely operated vehicle with wave disturbance preview,'' {\em The International Journal of Robotics Research}, 2024.

\bibitem{WalkerIROS}
K.~L. Walker and F.~Giorgio-Serchi, ``Disturbance preview for non-linear model predictive trajectory tracking of underwater vehicles in wave dominated environments,'' in {\em IEEE/RSJ International Conference on Intelligent Robots and Systems (IROS)}, (Detroit, MI, USA), pp.~6169--6176, 01-05 Oct 2024.

\bibitem{Tugal2023}
H.~Tugal, K.~Cetin, Y.~Petillot, M.~Dunnigan, and M.~S. Erden, ``Contact-based object inspection with mobile manipulators at near-optimal base locations,'' {\em Robotics and Autonomous Systems}, vol.~161, 2023.

\bibitem{Tonapi2015}
M.~M. Tonapi, I.~S. Godage, A.~M. Vijaykumar, and I.~D. Walker, ``A novel continuum robotic cable aimed at applications in space,'' {\em Advanced Robotics}, vol.~29, no.~13, pp.~861--875, 2015.

\bibitem{Qu2024Recent}
J.~Qu, Y.~Xu, Z.~Li, Z.~Yu, B.~Mao, Y.~Wang, Z.~Wang, Q.~Fan, X.~Qian, M.~Zhang, M.~Xu, B.~Liang, H.~Liu, X.~Wang, X.~Wang, and T.~Li, ``Recent advances on underwater soft robots,'' {\em Advanced Intelligent Systems}, vol.~6, no.~2, 2024.

\bibitem{Liu2020}
J.~Liu, S.~Iacoponi, C.~Laschi, L.~Wen, and M.~Calisti, ``Underwater mobile manipulation: A soft arm on a benthic legged robot,'' {\em IEEE Robotics \& Automation Magazine}, vol.~27, no.~4, pp.~12--26, 2020.

\bibitem{Nguyen2015}
T.-D. Nguyen and J.~Burgner-Kahrs, ``A tendon-driven continuum robot with extensible sections,'' in {\em IEEE/RSJ International Conference on Intelligent Robots and Systems (IROS)}, (Hamburg, Germany), pp.~2130--2135, 28 Sept - 02 Oct 2015.

\bibitem{Wockenfub2022}
W.~R. Wockenfu{\ss}, V.~Brandt, L.~Weisheit, and W.-G. Drossel, ``Design, modeling and validation of a tendon-driven soft continuum robot for planar motion based on variable stiffness structures,'' {\em IEEE Robotics and Automation Letters}, vol.~7, pp.~3985--3991, April 2022.

\bibitem{Xie2020}
Z.~Xie, F.~Yuan, Z.~Liu, Z.~Sun, E.~M. Knubben, and L.~Wen, ``A proprioceptive soft tentacle gripper based on crosswise stretchable sensors,'' {\em IEEE/ASME Transactions on Mechatronics}, vol.~25, no.~4, pp.~1841--1850, 2020.

\bibitem{guan2023trimmed}
Q.~Guan, F.~Stella, C.~Della~Santina, J.~Leng, and J.~Hughes, ``Trimmed helicoids: an architectured soft structure yielding soft robots with high precision, large workspace, and compliant interactions,'' {\em npj Robotics}, vol.~1, no.~1, p.~4, 2023.

\bibitem{mukaide2020radial}
R.~Mukaide, M.~Watanabe, K.~Tadakuma, Y.~Ozawa, T.~Takahashi, M.~Konyo, and S.~Tadokoro, ``Radial-layer jamming mechanism for string configuration,'' {\em IEEE Robotics and Automation Letters}, vol.~5, no.~4, pp.~5221--5228, 2020.

\bibitem{Wang2021}
M.~Wang, X.~Dong, W.~Ba, A.~Mohammad, D.~Axinte, and A.~Norton, ``Design, modelling and validation of a novel extra slender continuum robot for in-situ inspection and repair in aeroengine,'' {\em Robotics and Computer Integrated Manufacturing}, vol.~67, 2021.

\bibitem{Wang2018}
M.~Wang, D.~Palmer, X.~Dong, D.~Alatorre, D.~Axinte, and A.~Norton, ``Design and development of a slender dual-structure continuum robot for in-situ aeroengine repair,'' in {\em IEEE/RSJ International Conference on Intelligent Robots and Systems (IROS)}, (Madrid, Spain), pp.~5648--5653, 01-05 Oct 2018.

\bibitem{Dong2017}
X.~Dong, D.~Axinte, D.~Palmer, S.~Cobos, M.~Raffles, A.~Rabani, and J.~Kell, ``Development of a slender continuum robotic system for on-wing inspection/repair of gas turbine engines,'' {\em Robotics and Computer-Integrated Manufacturing}, vol.~44, pp.~218--229, 2017.

\bibitem{walker2024modular}
K.~L. Walker, A.~J. Partridge, H.-Y. Chen, R.~R. Ramachandran, A.~A. Stokes, K.~Tadakuma, L.~C. da~Silva, and F.~Giorgio-Serchi, ``A modular, tendon driven variable stiffness manipulator with internal routing for improved stability and increased payload capacity,'' {\em IEEE International Conference on Robotics and Automation (ICRA)}, pp.~3030--3035, 13-17 May 2024.

\bibitem{Webster_Jones_2010}
R.~J. Webster and B.~A. Jones, ``Design and kinematic modeling of constant curvature continuum robots: A review,'' {\em The International Journal of Robotics Research}, vol.~29, p.~1661–1683, Jun 2010.

\bibitem{Rucker_Webster_III_2011}
D.~C. Rucker and R.~J. Webster~III, ``Statics and dynamics of continuum robots with general tendon routing and external loading,'' {\em IEEE Transactions on Robotics}, vol.~27, p.~1033–1044, Dec 2011.

\bibitem{LynchBook}
K.~M. Lynch and F.~C. Park, {\em Modern Robotics: Mechanics, Planning, and Control}.
\newblock Cambridge University Press, 2017.

\bibitem{WalkerRoboSoft}
K.~L. Walker, C.~D. Santina, and F.~Giorgio-Serchi, ``Model predictive wave disturbance rejection for underwater soft robotic manipulators,'' {\em IEEE 7th International Conference on Soft Robotics (RoboSoft)}, pp.~40--47, 14-17 April 2024.

\end{thebibliography}

\end{document}